\documentclass{article}

\usepackage{PRIMEarxiv}

\usepackage[utf8]{inputenc} 
\usepackage[T1]{fontenc}    
\usepackage{hyperref}       
\usepackage{url}            
\usepackage{booktabs}       
\usepackage{amsfonts}       
\usepackage{nicefrac}       
\usepackage{microtype}      
\usepackage{lipsum}
\usepackage{fancyhdr}       
\usepackage{graphicx}       
\graphicspath{{media/}}     
 \usepackage{amsmath}
\title{EEG-Driven Image Reconstruction with Saliency-Guided Diffusion Models
}

\author{
  Igor Abramov \\
  Ivannikov Institute for System Programming of the Russian Academy of Sciences \\
  Research Center for Trusted Artificial Intelligence \\
  Moscow, Russia \\
  AI Talent Hub, ITMO University \\
  Saint Petersburg, Russia\\
  \texttt{ig.abramov@innopolis.university} \\
   \And
  Ilya Makarov \\
  AIRI \\
  Moscow, Russia \\
  Ivannikov Institute for System Programming of the Russian Academy of Sciences \\
  Research Center for Trusted Artificial Intelligence \\
  Moscow, Russia \\
  AI Talent Hub, ITMO University \\
  Saint Petersburg, Russia\\
  \texttt{makarov@airi.net} \\
}

\begin{document}
\maketitle

\begin{abstract}
Existing EEG-driven image reconstruction methods often overlook spatial attention mechanisms, limiting fidelity and semantic coherence. To address this, we propose a dual-conditioning framework that combines EEG embeddings with spatial saliency maps to enhance image generation. Our approach leverages the Adaptive Thinking Mapper (ATM) for EEG feature extraction and fine-tunes Stable Diffusion 2.1 via Low-Rank Adaptation (LoRA) to align neural signals with visual semantics, while a ControlNet branch conditions generation on saliency maps for spatial control. Evaluated on THINGS-EEG, our method achieves a significant improvement in the quality of low- and high-level image features over existing approaches. Simultaneously, strongly aligning with human visual attention. The results demonstrate that attentional priors resolve EEG ambiguities, enabling high-fidelity reconstructions with applications in medical diagnostics and neuroadaptive interfaces, advancing neural decoding through efficient adaptation of pre-trained diffusion models.
\end{abstract}

\keywords{Multimodal Interaction \and EEG \and Diffusion Models \and ControlNet \and LoRA \and Image Reconstruction \and BCI}

\section{Introduction}

The integration of neural data with generative models has emerged as a promising direction for brain-computer interfaces and cognitive computing. Within this domain, decoding visual experiences from brain activity remains a fundamental challenge \cite{Miyawaki2008,Jia2021}. While fMRI-based stimulus reconstruction shows promise \cite{Takagi2023,Ho2023}, EEG offers greater practicality through portability and temporal resolution \cite{Willett2021}. However, EEG's low signal-to-noise ratio has historically limited most work to classification rather than pixel-level synthesis.

Recent advances provide new pathways: The Adaptive Thinking Mapper (ATM) generates EEG embeddings aligned with visual semantics \cite{li2024visualdecodingreconstructioneeg}, while diffusion models \cite{latent-diffusion-rombach2021} and ControlNet \cite{controlnet-zhang2023adding} enable high-fidelity conditional image generation. Large-scale neuroimaging datasets (THINGS-EEG \cite{things-eeg-Gifford2022}, EEG-ImageNet \cite{eeg-imagenet}) have further accelerated progress. Crucially, existing approaches overlook visual attention patterns despite their perceptual importance \cite{NIPS2005_0738069b,NIPS2006_4db0f8b0,Itti2001} and potential for resolving EEG ambiguities \cite{zhang2024gazefusion}.

Saliency maps predict human visual attention patterns, offering interpretability for computer vision and conditioning signals for generative systems \cite{saliency-overview-Szczepankiewicz2023}. Early approaches used bottom-up statistical features \cite{NIPS2005_0738069b,NIPS2006_4db0f8b0,Itti2001}, while modern methods leverage deep learning and eye-tracking datasets (CAT2000 \cite{cat2000-dataset}, MIT1003 \cite{mit2003-dataset}, SALICON \cite{salicon-dataset}) to predict low-level and semantic attention \cite{deep-gaze-3-Kmmerer2022,salgan,eml-saliency-prediction-JIA20EML}. GazeFusion \cite{zhang2024gazefusion} demonstrated saliency-guided diffusion models, reflecting how attention combines low-level features and high-level semantics \cite{Kmmerer2015}.

\begin{figure}[h!bt]
\centering
\includegraphics[width=0.8\linewidth]{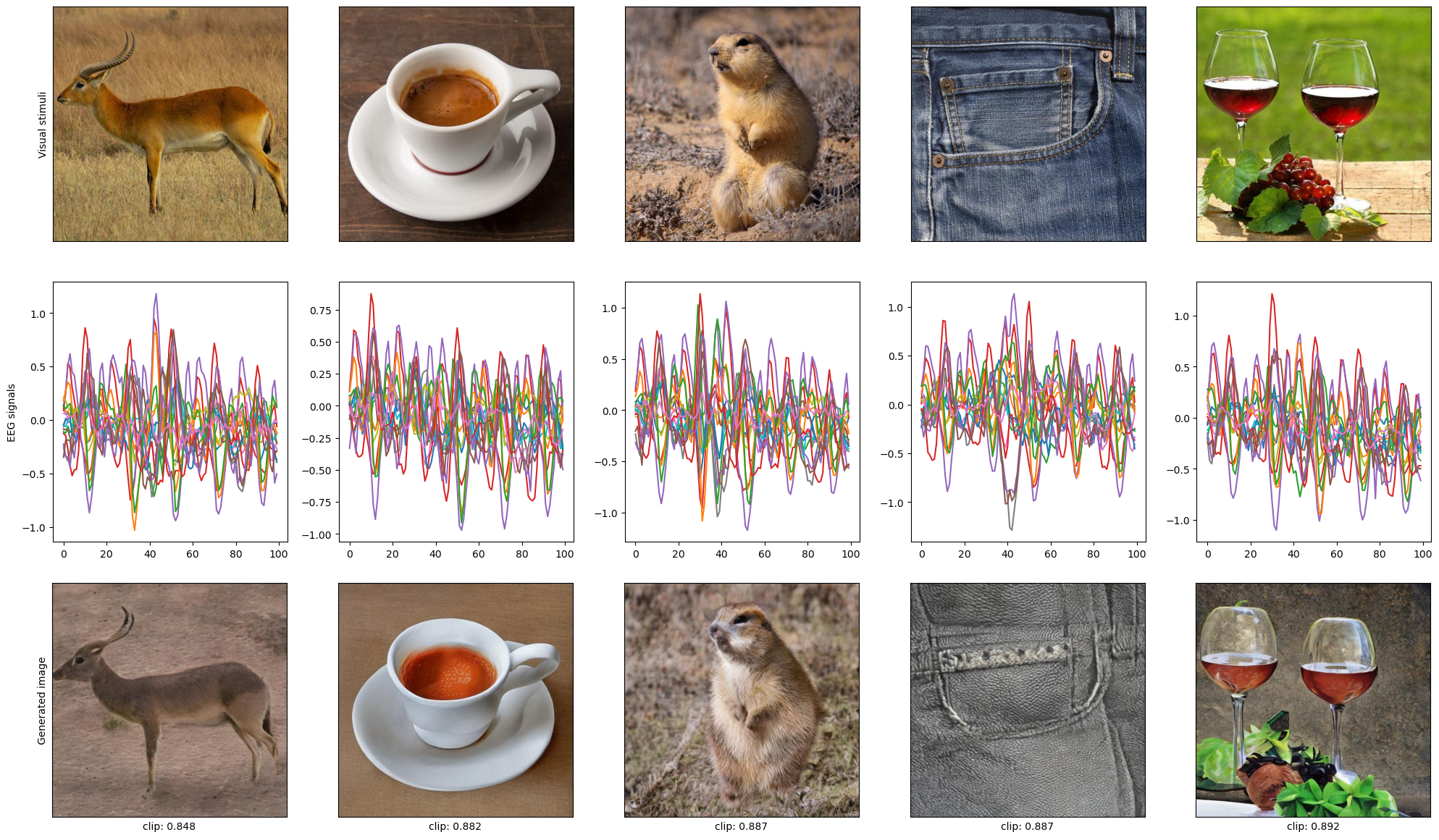}
\caption{Example stimuli and reconstructions showing original images (top), corresponding EEG signals (middle), and our model's outputs (bottom) conditioned on both EEG and saliency patterns.}
\label{fig:reconstructions}
\end{figure}

We bridge this gap with a novel dual-conditioning approach combining: 1) Semantic alignment of EEG embeddings via LoRA \cite{lora-hu2022}, and 2) Spatial guidance through ControlNet using predicted saliency maps. This integration leverages both semantic content and human-like attention patterns, significantly improving reconstruction quality (Fig. \ref{fig:reconstructions}) across pixel-level, structural, and semantic metrics versus EEG-only baselines. Implementation details are available at GitHub\footnote{\url{https://github.com/IGragon/EEG-Salience-Image}}.

\section{Proposed Framework}

\begin{figure}[h!bt]
\centering
\includegraphics[width=0.825\linewidth]{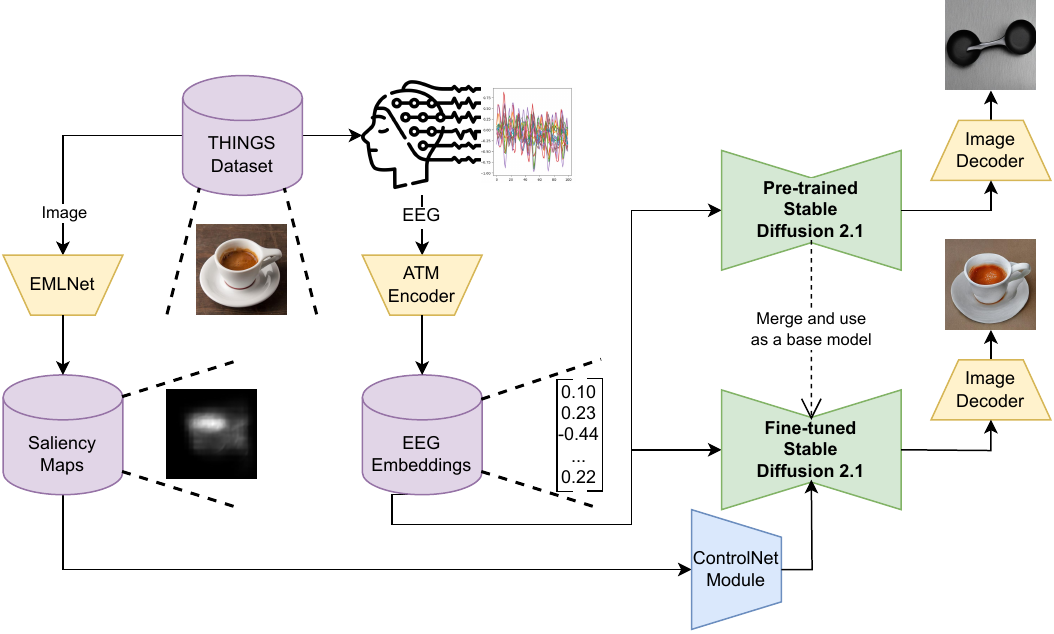}
\caption{Workflow of our EEG and saliency-conditioned image generation framework. Stage 1: LoRA fine-tuning of Stable Diffusion with EEG embeddings. Stage 2: ControlNet training with saliency maps while keeping the EEG-conditioned model frozen.}
\label{fig:workflow}
\end{figure}

Our framework integrates EEG-conditioned image generation with spatial attention control through a novel dual-conditioning approach, as illustrated in Figure~\ref{fig:workflow}. The system combines EEG feature extraction, latent diffusion modeling, and saliency-guided control in a multi-stage pipeline.

We employ the Adaptive Thinking Mapper (ATM) encoder \cite{li2024visualdecodingreconstructioneeg} to process raw EEG signals from the THINGS-EEG dataset \cite{things-eeg-Gifford2022}. The architecture preserves EEG's spatial and temporal features using channel-wise attention and specialized convolutions, producing meaningful embeddings that match visual concepts.

Building on Stable Diffusion 2.1 \cite{latent-diffusion-rombach2021}, we adapt the image generation process to accept EEG embeddings as semantic conditioning. Using Low-Rank Adaptation (LoRA) \cite{lora-hu2022}, we efficiently fine-tune the cross-attention layers of the UNet to respond to neural patterns while preserving the model's original generative capabilities. We trained this setup on single RTX4080 in half precision. AdamW \cite{adamw} with $\text{lr}=2e^{-3}$ and default betas performed 212000 optimization steps with batch size 8. We also used Cosine Annealing Warm Restarts \cite{cosine-annealing-warm-restarts} lr scheduler with $\text{eta\_min}=1e^{-6}$, $T_0=5000$.

The final enhancement incorporates ControlNet \cite{controlnet-zhang2023adding} to condition generation on predicted saliency maps from EMLNet \cite{eml-saliency-prediction-JIA20EML}. This spatial conditioning branch directs image composition according to human attention patterns while maintaining the semantic content derived from EEG. We trained this setup on single RTX4080 in full presicion. AdamW \cite{adamw} with $\text{lr}=1e^{-4}$ and default betas performed 65000 optimization steps with batch size 2 and 4 gradient accumulation steps. We also used Cosine Annealing Warm Restarts \cite{cosine-annealing-warm-restarts} lr scheduler with $\text{eta\_min}=1e^{-5}$, $T_0=5000$.

As shown in Figure~\ref{fig:reconstructions}, the combined conditioning produces images that align with both the conceptual and attentional aspects of human vision.

\section{Experimental Results}
\label{sec:experimental-results}

We evaluate our dual-conditioning approach using reconstruction metrics (low-level and semantic) and saliency metrics. Quantitative comparisons use Subject 8 data from THINGS-EEG \cite{things-eeg-Gifford2022} against EEG-based image reconstruction framework proposed in \cite{li2024visualdecodingreconstructioneeg} with different EEG encoders.

Tables \ref{tab:image-metrics-low-level} and \ref{tab:image-metrics-high-level} demonstrate significant improvements. Our saliency-guided approach achieves a significantly higher improvement in pixel correlation (PixCorr) and structure similarity index (SSIM) over the \cite{li2024visualdecodingreconstructioneeg} framework. Semantic metrics (i.e. \cite{alexnet, inception-v3, clip-radford2021, swav} using \cite{Ozcelik2023}) show near-perfect scores, confirming superior preservation of both low-level features and high-level semantics.

\begin{table}[h!]
\centering
\caption{Low-level reconstruction metrics (Subject 8). Higher is better.}
\label{tab:image-metrics-low-level}
\begin{tabular}{@{}lcc@{}}
\toprule
Approach & PixCorr $\uparrow$ & SSIM $\uparrow$ \\ \midrule
VDaR, ATM \cite{li2024visualdecodingreconstructioneeg} & 0.160 & 0.345 \\
Ours: EEG-only & 0.080 & 0.271 \\
Ours: Saliency-guided & \textbf{0.473} & \textbf{0.369} \\ \bottomrule
\end{tabular}
\end{table}

\begin{table}[h!]
\centering
\caption{High-level reconstruction metrics (Subject 8).}
\label{tab:image-metrics-high-level}

\begin{tabular}{@{}lccccc@{}}
\toprule
Approach & \multicolumn{1}{l}{AlexNet (2) $\uparrow$} & \multicolumn{1}{l}{AlexNet (5) $\uparrow$} & \multicolumn{1}{l}{Inception $\uparrow$} & \multicolumn{1}{l}{CLIP $\uparrow$} & \multicolumn{1}{l}{SwAV $\downarrow$} \\ \midrule
VDaR, ATM \cite{li2024visualdecodingreconstructioneeg} & 0.776 & 0.866 & 0.734 & 0.786 & 0.582 \\
Ours: EEG-only & 0.774      & 0.865                          & 0.745                         & 0.767                    & 0.593 \\
Ours: Saliency-guided & \textbf{0.999} & \textbf{0.998} & \textbf{0.946} & \textbf{0.904} & \textbf{0.453} \\ \bottomrule
\end{tabular}%

\end{table}

Table \ref{tab:saliency-metrics} confirms our spatial conditioning significantly improves attention alignment. Saliency control yields higher correlation coefficient (CC), lower KL divergence, and higher SIM compared to EEG-only conditioning.

\begin{table}[h!]
\centering
\caption{Saliency metrics (Subject 8).}
\label{tab:saliency-metrics}
\begin{tabular}{@{}lccc@{}}
\toprule
Approach & CC $\uparrow$ & KL $\downarrow$ & SIM $\uparrow$ \\ \midrule
EEG-only & 0.51 & 2.99 & 0.60 \\
Saliency-guided & \textbf{0.85} & \textbf{0.52} & \textbf{0.80} \\ \bottomrule
\end{tabular}
\end{table}

\section{Conclusion}

We presented a novel EEG-conditioned image generation framework enhanced with saliency guidance, demonstrating significant improvements in both reconstruction quality and attention alignment. Our approach achieves noticeable performance with higher pixel correlation and better saliency correlation compared to EEG-only baselines, validating that spatial attention cues resolve ambiguities in neural decoding. The work establishes that parameter-efficient adaptation of diffusion models can effectively incorporate both EEG embeddings and saliency maps, opening new possibilities for brain-computer interfaces. Future directions include EEG-predicted saliency estimation and cross-subject generalization to advance practical applications in cognitive neuroscience and assistive technologies.

\section*{Acknowledgments}
The work was supported by a grant,
provided by the Ministry of Economic Development of the Russian
Federation in accordance with the subsidy agreement (agreement
identifier 000000C313925P4G0002) and the agreement with the Ivannikov Institute for System Programming of the Russian Academy
of Sciences dated June 20, 2025 No. 139-15-2025-011.

\bibliographystyle{unsrt}  
\bibliography{references}

\end{document}